\documentclass[11pt,a4paper]{article}
\usepackage[hyperref]{acl2020}
\usepackage{times}
\usepackage{latexsym}

\usepackage{microtype}

\aclfinalcopy 
\def\aclpaperid{962} 

\usepackage{amsmath}
\usepackage{amssymb}

\usepackage{booktabs} 
\usepackage{multirow}
\usepackage{enumitem}
\usepackage{hhline}
\usepackage{pgf}
\usepackage{lscape} 

\usepackage{array}
\newcolumntype{L}[1]{>{\raggedright\let\newline\\\arraybackslash\hspace{0pt}}p{#1}}
\newcolumntype{C}[1]{>{\centering\let\newline\\\arraybackslash\hspace{0pt}}m{#1}}
\newcolumntype{R}[1]{>{\raggedleft\let\newline\\\arraybackslash\hspace{0pt}}m{#1}}

\usepackage{graphicx}
\usepackage{multirow}
\usepackage{caption}
\usepackage{subcaption}
\usepackage[bottom]{footmisc} 
\usepackage{dblfloatfix}  
\usepackage{booktabs}
\usepackage[capposition=bottom]{floatrow} 

\title{Better Early than Late: Fusing Topics with Word Embeddings\\ for Neural Question Paraphrase Identification}
\author{Nicole Peinelt$^{1,2}$  \and  Dong Nguyen$^{1,3}$ \and Maria Liakata$^{1,2}$\\
$^1$The Alan Turing Institute, London, UK \\
$^2$University of Warwick, Coventry, UK \\
$^3$Utrecht University, Utrecht, The Netherlands \\
  {\tt \{n.peinelt, m.liakata\}@warwick.ac.uk}, {\tt dnguyen@turing.ac.uk}}

\date{}

\begin{document}
\maketitle
\begin{abstract}
Question paraphrase identification is a key task in Community Question Answering (CQA) to determine if an incoming question has been previously asked. Many current models use word embeddings to identify duplicate questions, but the use of topic models in feature-engineered systems suggests that they can be helpful for this task, too. We therefore propose two ways of merging topics with word embeddings (early vs. late fusion) in a new neural architecture for question paraphrase identification. Our results show that our system outperforms neural baselines on multiple CQA 
datasets, while an ablation study highlights the importance of topics and especially \emph{early} topic-embedding fusion in our architecture.
\end{abstract}

\section{Introduction}
\label{intro}

Paraphrase identification is a core NLP task and has been widely studied \cite{socher_dynamic_2011,he_multi-perspective_2015,wieting_towards_2016,tomar_neural_2017}.
One interesting application area of paraphrase detection is Community Question Answering (CQA) \cite{SemEval-2017:task3,bonadiman_effective_2017,rodrigues_semantic_2018}. 
The aim of CQA is to answer real open-ended questions based on user-generated content from question answering websites. Being able to identify similar --- already answered --- questions 
can be helpful for this purpose.
Question paraphrase detection in CQA is difficult
because texts tend to be longer and have less direct overlap compared to traditional paraphrase detection datasets \cite{rus_paraphrase_2014, peinelt_aiming_2019}.

Early work on paraphrase detection relied on hand-crafted features, while state-of-the-art approaches for paraphrase identification are primarily neural networks \cite{gong_natural_2018,wang_bilateral_2017,tomar_neural_2017} and hybrid techniques \cite{pang_text_2016,wu_ecnu_2017,feng_beihang-msra_2017}.
Many recently proposed CQA paraphrase detection systems still use hand-crafted features
\cite{agustian_uinsuska-titech_2017,filice_kelp_2017} and some work has successfully integrated topic model features \cite{duan_searching_2008,wu_ecnu_2017}.
This suggests that topic distributions could offer auxiliary information for identifying related questions and complement word embeddings \cite{mikolov_efficient_2013,pennington_glove:_2014}, which provide the main signal in neural systems. 
Contrary to hand-crafted static topic features, integrating topics in a neural framework brings the advantage of joint updates during training. 
Recent work successfully introduced topics in neural architectures for language generation:
\citet{wang_reinforced_2018} used a topic-enhanced encoder for summarisation, \citet{chen_guided_2016} integrated topics in the decoder for machine translation and \citet{narayan_dont_2018} included topics in both encoder and decoder of their summarisation model.

However, it remains unclear if topics can be useful in a neural paraphrase detection model and how to best fuse topics with word embeddings for this task. 
In this paper, we introduce a novel topic-aware neural architecture and specifically make the following contributions:
\begin{enumerate}
\item We define two settings (early and late fusion) for incorporating topics in our neural paraphrase prediction model (section \ref{sec:model}).
\item Our topic-aware model improves over other neural models across multiple question paraphrase identification datasets (section \ref{sec:results}).
\item In an ablation study, we highlight the importance of topics and early topic-embedding fusion in our proposed architecture (section \ref{sec:results}).
\end{enumerate}

\section{Datasets and Tasks}
\label{sec:tasks}

We address the problem of CQA question paraphrase detection, where given two questions from question answering websites, denoted as $q_1$ and $q_2$ with length $n$ and $m$, the task is to predict a binary label which indicates whether the two questions are paraphrases.
For this study, we select three popular English  question paraphrase identification datasets and summarise their main properties in Table~\ref{tab:cQA_datasets}.

\begin{table}[t]
  \small
  \centering
    \begin{tabular}{lllr}
    \toprule
 & Task  &  Source  & Question pairs  \\ 
    	\midrule
Quora  & PD  & real-world  & 404 345   \\
PAWS & PD & synthetic  & 12 665\\
SemEval & PR   & real-world  & 4 749  \\
    \bottomrule
    \end{tabular}%
   \caption{Selected CQA question paraphrase datasets. PD=paraphrase detection, PR=paraphrase ranking}
       \label{tab:cQA_datasets}
\end{table}%

The \textbf{Quora} duplicate questions dataset 
consists of over 400 000 question pairs, each annotated with a binary label indicating whether the two questions are paraphrases of each other or not.\footnote{https://engineering.quora.com/Semantic-Question-Matching-with-Deep-Learning} We use the same split as mentioned in \citet{wang_bilateral_2017}.

\begin{figure}[b]
  \centering
  \includegraphics[height=6.1cm]{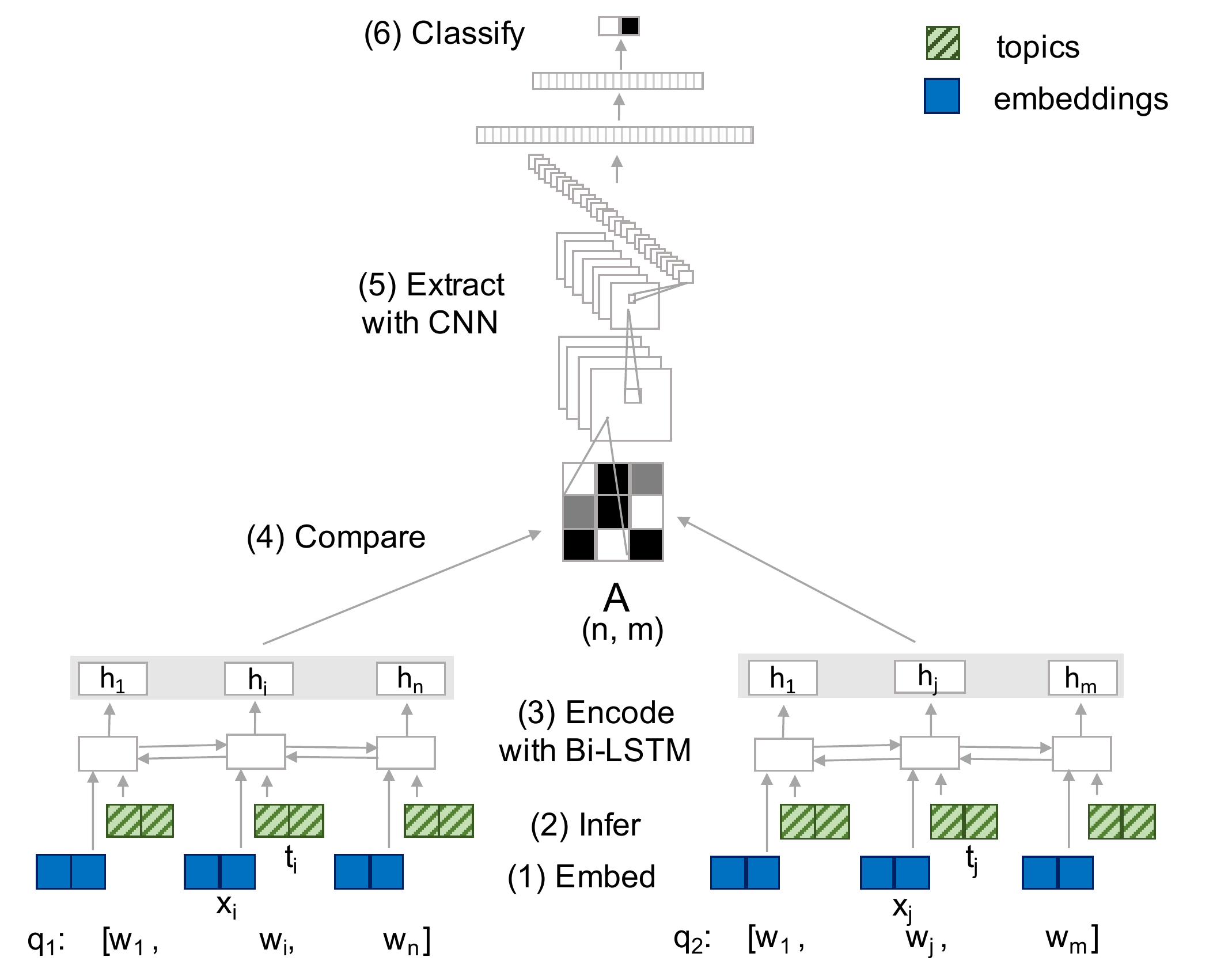}
  \caption{TAPA with early topic-embedding fusion. Illustrated with toy examples consisting of three tokens.}
  \label{fig_model_comb}
\end{figure}

\begin{figure}[b]
  \centering
  \includegraphics[height=6.1cm]{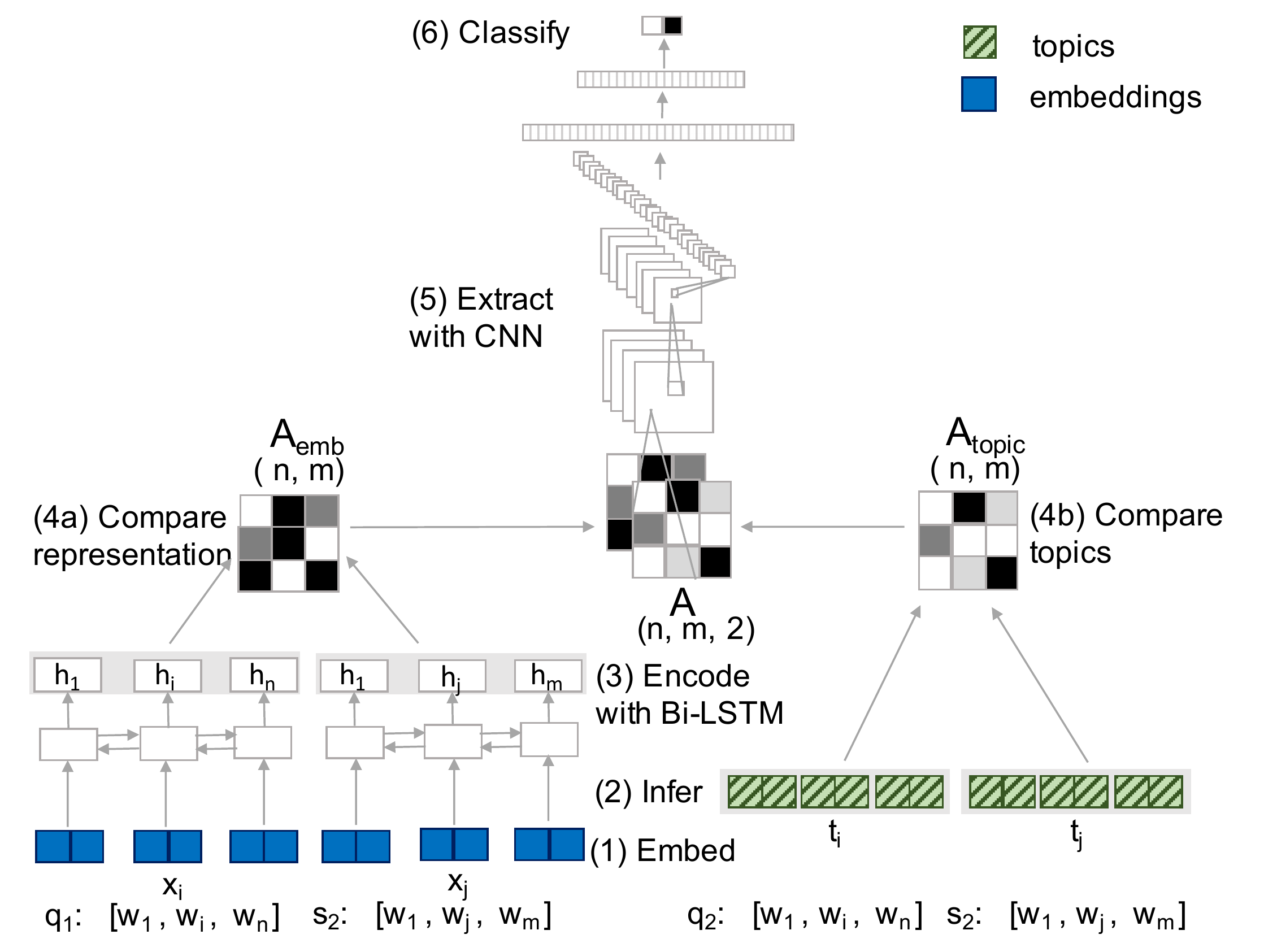}
  \caption{TAPA with late topic-embedding fusion}
  \label{fig_model_sep}
\end{figure}

\textbf{PAWS} \cite{zhang_paws_2019} is a synthetic paraphrase detection dataset which was created on the basis of Quora. The word order of Quora examples was automatically altered and words replaced, resulting in question pairs with high word overlap which were manually annotated with binary paraphrase labels.

The \textbf{SemEval} 2017 Task 3 dataset focuses on Community Question Answering \cite{SemEval-2017:task3}. While there are other subtasks, we only use subtask B (question paraphrase detection) here.
In each example, the dataset provides a new question and a set of ten possibly related questions which were retrieved from the forum of the website \emph{Qatar Living}\footnote{https://www.qatarliving.com/} by a search engine. For each question pair, a relevancy label is provided.

\section{Model Architecture}
\label{sec:model}

This paper examines whether we can successfully integrate topic model information in a neural paraphrase identification model. We also explore how to best combine topics with word embeddings, which are the main information source in existing models \cite{deriu_swissalps_2017,pang_text_2016,gong_natural_2018}.
For this purpose, we propose a novel architecture dubbed \textbf{T}opic-\textbf{A}ware \textbf{P}araphrase Identification \textbf{A}rchitecture (TAPA) depicted in Figures \ref{fig_model_comb} and \ref{fig_model_sep}.
Our model comprises the following steps:
Obtaining a representation for both questions (Section \ref{sec:model_encoder}), comparing these representations (Section \ref{sec:model_comparison}), and aggregating the information for the final prediction (Section \ref{sec:model_comparison}).

\subsection{Encoding layer}
\label{sec:model_encoder}

\paragraph{Embeddings} 
Given the sequence of words $(w_1,...,w_n)$ in a question of length $n$, we map them to embeddings $x = (x_1,...,x_n)$ (step 1 in Figures \ref{fig_model_comb} and \ref{fig_model_sep}).
For each embedding $x_i$, we combine pretrained word 
and ELMo embeddings \cite{peters_deep_2018-1} to leverage alternative representations:
\begin{align}
x_i = [\text{emb}_i; \text{ELMo}_i] \in R^{f}
\end{align}
where ; denotes concatenation and $f$ the resulting embedding dimension.
\paragraph{Topics} 
We infer topic distributions from an LDA topic model \cite{blei_latent_2003} with  $f'$ number of topics for the whole question $t_D \in R^{f'}$ and every word in the question $t' = (t'_1,...,t'_n)$ with $t_i \in R^{f'}$ (step 2 in Figures \ref{fig_model_comb} and \ref{fig_model_sep}).
We follow \citet{narayan_dont_2018} in merging word-level and document-level topics to capture the interaction between both as 
\begin{align}
t_i = [t'_i \otimes t_D] \in R^{f'}
\end{align}
where $\otimes$ denotes element-wise multiplication as `word+doc' setting.
Previous work by \citet{narayan_dont_2018} focused on a different task (summarisation) with longer texts. In contrast, for our setting we expect word-level topics to be more informative than document topics.
Therefore, we further include only word-level topics as `word' setting
\begin{align}
t_i = [t'_i] \in R^{f'}
\end{align}
where the topic setting 
is treated as hyperparameter.

\paragraph{Fusion of embeddings and topics} 
We propose two different ways of combining topic distributions with word embeddings: early fusion and late fusion.
Early fusion combines topic distributions with word embeddings before the encoder step (Figure \ref{fig_model_comb}), guiding the encoder in selecting relevant information when computing a representation for the sentence.
In contrast, late fusion combines information derived from topics and word embeddings \emph{after} computing separate affinity matrices for topics and word representations (Figure \ref{fig_model_sep}), therefore introducing the topic information more directly into the architecture as separate sentence interaction dimension. As a result, the encoding layer of both variations differs slightly. In early fusion TAPA, we obtain a sentence representation  $\text{\bf{e}} = (e_1,...,e_n)$ by concatenating topics and embeddings as follows:
\begin{align}
 e_i = [x_i; t_i] \in R^{f+f'}
 \end{align}
Late fusion TAPA only uses embeddings for $\text{\bf{e}}$
\begin{align}
 e_i = [x_i] \in R^{f}
 \end{align}
 and obtains separate topic representations 
 \begin{align}
  \text{TL} = [t_1,...,t_n] \in R^{(n \times f')}\\
  \text{TR} = [t_1,...,t_m] \in R^{(m \times f')}
\end{align}

\paragraph{Encoder}
In our preliminary experiments, BiLSTMs \cite{hochreiter_long_1997} worked better than CNN encoders, presumably due to their ability to capture long-range dependencies.
As a result, we encode $\bf{e}$ with BiLSTMs  (step 3 in Figures \ref{fig_model_comb} and \ref{fig_model_sep}):
\begin{align}
\text{L} = \text{BiLSTM}(\text{\bf{e}} ) \in R^{(n\times d)}\\
\text{R} = \text{BiLSTM}(\text{\bf{e}} ) \in R^{(m\times d)}
\end{align}
We decide to share weights between both BiLSTMs in a Siamese setting which reduces the number of required parameters and has been shown to work well for pairwise classification tasks \cite{deriu_swissalps_2017,feng_applying_2015,mueller_siamese_2016}.

\subsection{Comparison layer}
\label{sec:model_comparison}

We model the similarity between the two encoded questions by computing pairwise affinity scores between their words in affinity matrices similar to previous studies \cite{deriu_swissalps_2017,pang_text_2016,gong_natural_2018}.
In both early and late fusion TAPA, we calculate the affinity matrix $A_{emb}$ based on the encoded sentences as 
\begin{align}
A_{emb_{i,j}} = s(\text{L}_{i,:},\text{R}_{j,:}) \in \mathbb{R}^{n \times m}
\end{align}
using a similarity function $s$ (step 4 in Figure \ref{fig_model_comb} and 4a in Figure \ref{fig_model_sep}). 
For the late fusion model, we also calculate an affinity matrix based on the topic distributions in both sentences (4b in Figure \ref{fig_model_sep}):\footnote{We experimented with additional topic encoders, but abandoned them as they didn't consistently improve results.}
\begin{align}
A_{topic_{i,j}} = s(\text{TL}_{i,:},\text{TR}_{j,:})  \in \mathbb{R}^{n \times m}
\end{align}
 Common choices for the similarity function $s$ are Euclidean distance, dot product and cosine similarity \cite{yin_abcnn:_2016,deriu_swissalps_2017}. We use cosine similarity, as it  worked best in preliminary experiments.
The comparison layer output $A$ is simply $A_{emb}$ for early fusion TAPA, while we combine topic and embedding affinity matrices in the late fusion version:
$$A=[A_{emb};A_{topic}]$$

\subsection{Aggregation layer}
\label{sec:model_aggregation}

Similar to extracting information from a grey scale image in computer vision, we
follow \citet{gong_natural_2018} and \citet{pang_text_2016} in aggregating useful affinity patterns from $A$ with a CNN  \cite{lecun_gradient-based_1998} feature extractor (step 5 in in Figure \ref{fig_model_comb} and \ref{fig_model_sep}). We use a two-layer CNN architecture (where one layer consists of convolution and pooling). The output of the last convolution layer is flattened into a vector.
This is followed by multiple hidden layers of reducing size and a softmax layer for predicting the two classes (step 6 in Figure \ref{fig_model_comb} and \ref{fig_model_sep}). The model is trained based on cross-entropy loss. We tune hyperparameters with hyperopt \cite{bergstra_hyperopt_2013} and report them in Appendix B. For further implementation details refer to Appendix C.



\begin{table}[b]
  \small
  \centering
\begin{tabular}{L{2.11820225cm}L{0.79045868cm}R{0.8215cm}R{0.79047cm}R{1.134285cm}}
\toprule
  & Type &PAWS & Quora & SemEval\\ 
\midrule
KeLP  & feat. & - & - & \bf{50.6} \\ 
NLM-NIH  & feat. & - & - & 47.3 \\ 
Uinsuska TiTech & feat. & - & - & 46.7 \\ 
Siamese  network    &  neural &   17.3 &     81.3 &        34.9 \\
+ELMo &    neural &  37.2 &  83.2 &     34.5 \\
TAPA     & neural &   \bf{42.2} &     \bf{84.1} &     46.4 \\
\bottomrule
    \end{tabular}%
  \caption{F1 scores of models on test sets. The first three rows are taken from \citet{SemEval-2017:task3}, the rest are our own implementations. feat=feature-based}
      \label{tab:results}
\end{table}%

\begin{table}[t]
  \small
  \centering
\begin{tabular}{lrrR{1cm}}
\toprule
  & PAWS & Quora & Sem-Eval\\ 
\midrule
full TAPA (early fusion) &  42.2 &    84.1 &    46.4 \\
-topics & 40.6 & 83.9 & 45.1 \\
-ELMO & 26.9 & 84.5 & 45.0\\
TAPA with late fusion & 39.8 & 83.9 & 40.1 \\
\bottomrule
    \end{tabular}%
  \caption{Ablation study for our TAPA model reporting F1 scores on test sets.}
      \label{tab:ablation}
\end{table}%

\section{Results}
\label{sec:results}

We evaluate model performance on the basis of F1 scores as this is more reliable for datasets with label imbalance than accuracy and present the results in Table~\ref{tab:results}. 

\paragraph{Baselines}
As baseline systems, we provide the three best performing \textbf{SemEval 2017 models}: KeLP \cite{filice_kelp_2017}, NLM-NIH \cite{abacha_nlm_2017} and Uinsuska TiTech \cite{agustian_uinsuska-titech_2017}.  All of these models employ hand-crafted features, which provides an advantage on this small dataset (compare Table~\ref{tab:cQA_datasets} for dataset sizes). 
Results have been reported for systems on PAWS and Quora with accuracies ranging between 75 and 89, but are not directly comparable to F1 scores \cite{gong_natural_2018,tan_multiway_2018,tomar_neural_2017}.

In addition to the above mentioned systems, we also compare our proposed model with a
 \textbf{Siamese network}, as this is a common neural baseline for paraphrase identification \cite{wang_bilateral_2017}. The two questions are embedded with pretrained 300 dimensional Glove embeddings \cite{pennington_glove:_2014} and encoded by two weight-sharing BiLSTM encoders. This is followed by a max pooling layer and two hidden layers. For Siamese network +ELMo, we concatenate word vectors with ELMo representations before the encoding step.

\paragraph{TAPA vs. baselines}
TAPA performs better than the neural baselines.
It cannot compete with the highest ranked feature engineered SemEval system (KeLP) due to the lack of sufficient training data (compare Table \ref{tab:cQA_datasets}), but gets within reach of the third placed system (Uinsuska TiTech). Our neural architecture may not outperform the top three systems on the SemEval dataset, but it can generalise better across datasets, while the three SemEval systems require dataset specific feature engineering.

\paragraph{Influence of model components} 
We conduct an ablation study to understand the contribution of individual model components (Table \ref{tab:ablation}). Removing topics consistently reduces F1 scores on all datasets, while the effect of ELMo representations is dataset dependent. Deleting ELMo improves performance on Quora, but leads to a massive performance drop on PAWS. The large impact on PAWS can be explained by the fact that this dataset was automatically constructed to have high textual overlap between questions and differences between paraphrases are chiefly due to variations in syntax.
Our full TAPA model uses early fusion as this was the best setting during hyperparameter tuning. When comparing the full (early fusion) model with a tuned\footnote{As late fusion may require slightly different hyperparameters, we compare with a tuned late fusion model to make a fair comparison between early and late fusion.} late fusion variant, we find that performance of the late fusion model drops to the same or even lower level than TAPA without topics. We conclude that topics contribute consistently to the performance of our proposed model, but that early topic-embedding fusion is crucial.

\section{Conclusion}
\label{conclusion}

In this work, we introduced a novel topic-aware neural architecture for question paraphrase identification. Our model successfully fuses word embeddings with topics and improves over previous neural baselines on multiple CQA paraphrase identification datasets.
We demonstrated that topics contributed consistently to the performance of our model and that an early fusion of word embeddings with topic distributions is preferable over integration at a later stage. 
Our work suggests that early fusion of topics with
models which were pretrained with sentence pair classification tasks, such as BERT \cite{devlin_bert_2019} could be a promising direction for future research.
Other future work could seek to enhance our proposed architecture with more sophisticated topic models.

\bibliography{notes}
\bibliographystyle{acl_natbib}

\onecolumn

\section*{Appendix A: Examples}
\label{appendix:examples}

\begin{table}[h]
  \small
  \centering
        \begin{tabular}{llL{0.72025cm}}
        \toprule
      Dataset  &    Sentence pair & Label \\
        \midrule
        \multirow{ 2}{*}{Quora} &  Which is the best way to learn coding? & \multirow{ 2}{*}{1} \\
                &       How do you learn to program?  & \\
        \midrule
        \multirow{ 2}{*}{PAWS} & How is Hillary Clinton a better choice than Donald Trump? & \multirow{ 2}{*}{0} \\
            &   How is Donald Trump a better choice than Hillary Clinton?      & \\
        \midrule
        \multirow{ 2}{*}{SemEval} & Where I can buy good oil for massage? & \multirow{ 2}{*}{1} \\
          &     Is there any place i can find scented massage oils in Qatar?  & \\
        \bottomrule
        \end{tabular}
    \caption{Examples from different datasets.}
     \label{tab:difficulty_examples}
\end{table}
 
 \section*{Appendix B: Hyperparameters}
\label{appendix:hyperparam}

PAWS does not have a development set for hyperparameter tuning and we only use its test set to evaluate our Quora model, matching one of the reported settings in Zhang et al. (2019). 

\begin{table*}[h]
  \small
  \centering
    \begin{tabular}{C{0.8cm}C{0.8379cm}C{0.6798cm}C{0.5cm}C{0.686cm}C{0.712cm}C{1.15cm}C{0.78cm}C{1.3cm}C{0.78cm}C{0.686cm}C{0.844cm}C{0.665cm}}
    \toprule
dataset &  \# of filters & filter size &  \# of hl & batch size & lr & optimizer & embd & topic type & \# of topics & topic alpha & topic update & fusion\\ 
\midrule
Quora &  (4, 12) & (2, 2) & 2 & 64 & 0.050 & adadelta & Glove & word & 70 & 50 & True& early\\ 
PAWS &  (4, 12) & (2, 2) & 2 & 64 & 0.050 & adadelta & Glove & word & 70 & 50 & True& early\\ 
Sem-Eval &  (0, 0) & (0, 0) & 2 & 10 & 0.100 & adadelta & Deriu & word & 90 & 0.1 & True& early\\
    \bottomrule
    \end{tabular}%
   \caption{Hyperparameters obtained after tuning on development set. hl=hidden layers, lr=learning rate, filters=filters of CNN feature extractor. Deriu embedding as used in Deriu and Cieliebak (2017).
   }
       \label{tab:hyperparam}
\end{table*}%

\section*{Appendix C: Implementation Details}
\label{appendix:implementation}

All words were lower-cased during preprocessing. Embeddings are initialised with pretrained 300 dim Glove embeddings 
\cite{pennington_glove:_2014}  
which were updated during training. 
We trained LDA topic models \cite{blei_latent_2003} 
on the training set of each dataset using \emph{ldamallet} from the Gensim package  \cite{rehurek_software_2010} 
and experimented with static vs. updated topic distributions, different alpha values (0.1 to 50) and number of topics (10 to 100) which are treated as hyperparameters. As PAWS is based on Quora, we used Quora topic models for both datasets.
We tune hyperparameters with the tree of Parzen estimators algorithm implemented in Hyperopt \cite{bergstra_hyperopt_2013} 
based on dev set F1, see Appendix B for our best hyperparameters. 

\end{document}